\documentclass[10pt,a4paper]{article}

\usepackage{lrec2026}
\usepackage{booktabs}
\usepackage[table]{xcolor}
\usepackage{tabularx}
\usepackage{amsmath}
\usepackage{amssymb}
\usepackage{comment}
\usepackage{enumitem}

\DeclareUnicodeCharacter{0101}{\={a}}
\DeclareUnicodeCharacter{012B}{\={i}}
\DeclareUnicodeCharacter{016B}{\={u}}
\DeclareUnicodeCharacter{02BE}{'}
\DeclareUnicodeCharacter{02BF}{`}
\DeclareUnicodeCharacter{1E0D}{\d{d}}
\DeclareUnicodeCharacter{1E25}{\d{h}}
\DeclareUnicodeCharacter{1E63}{\d{s}}
\DeclareUnicodeCharacter{1EB7}{\u{a}}
\DeclareUnicodeCharacter{2019}{'}

\title{QIAS 2026: Overview of the Shared Task on Islamic Inheritance Reasoning}
\name{\bf  Abdessalam BOUCHEKIF$^1$, Somaya ELTANBOULY$^1$, Samer RASHWANI$^1$, \\
      \bf Shahd GABEN$^1$, Mutaz AL-KHATIB$^1$, Heba SBAHI$^1$, \\
      \bf  Emad MOHAMED$^2$, Mohammed GHALY$^1$}

\address{$^1$Hamad bin Khalifa University, Qatar \quad
         $^2$Nazarbayev University, Kazakhstan \\
         \texttt{\{abouchekif,seltanbouly,srashwani,sgaben,malkhatib,hsbahi,mghaly\}@hbku.edu.qa} \\
         \texttt{emad.mohamed@nu.edu.kz}}
         
\abstract{
This paper presents a comprehensive overview of the QIAS 2026 shared task, organized as part of the OSACT7 Workshop and co-located with LREC 2026. The shared task was designed to evaluate the ability of large language models to perform complex reasoning in the religious and legal domain of Islamic inheritance. Unlike conventional question-answering benchmarks, QIAS 2026 focuses on end-to-end reasoning from natural language cases, requiring systems to perform the full inheritance calculation process, from identifying the eligible heirs to assigning the correct share to each beneficiary. To support this evaluation, the task was based on the MAWARITH benchmark, a dataset of $12{,}500$ Arabic inheritance cases annotated with intermediate reasoning steps and final answers. System submissions were evaluated using MIR-E, a multi-step metric that measures performance across the main stages of inheritance reasoning. A total of $16$ teams participated in the shared task, investigating a range of approaches, including prompting-based methods, retrieval-augmented generation, and fine-tuning strategies. The results show that Islamic inheritance remains a highly challenging benchmark for current language models, especially in stages that require precise legal interpretation and structured numerical reasoning. This overview summarizes the task design, dataset, evaluation framework, participating systems, and main results.
 \\ \newline  }

\begin{document}

\maketitleabstract
\section{Introduction} \label{intro}
Large language models (LLMs) have recently achieved strong performance across a wide range of natural language processing tasks, including question answering, summarization, and complex text generation. Their success has been particularly evident in tasks that benefit from broad linguistic coverage and large-scale pretraining. However, LLMs still face important challenges in specialized domains that require precise reasoning, structured decision-making, and strict adherence to domain-specific rules. These limitations become more apparent in tasks that require a sequence of dependent reasoning steps, where an error at an early stage can propagate and compromise the final answer. This issue is especially important in religious and legal domains, where reasoning is not only knowledge-intensive but also constrained by formal principles and interpretive traditions. 

In Islamic studies, and particularly in Islamic law, systems must reason over authoritative and highly structured sources such as the Qur'an, Hadith, and juristic writings. They must also operate within a framework based on clear legal principles and, in some cases, different interpretations across schools of thought. As a result, evaluating LLMs in such contexts requires benchmarks that go beyond surface-level question answering and instead test the ability of models to produce precise and faithful reasoning. Islamic inheritance law (\textit{ʿilm al-mawārīth}) serves as an effective testbed for evaluating reasoning capabilities. This domain of Islamic jurisprudence requires multi-step legal and numerical reasoning to resolve cases. A valid solution must identify eligible heirs, determine which heirs are excluded or blocked, assign appropriate shares, assess the need for adjustments, and calculate the final distribution. The process is governed by strict jurisprudential rules, and certain cases present additional complexities such as \textit{ʿawl} and \textit{radd}. 
Due to its combination of legal interpretation, structured reasoning, and precise calculation, Islamic inheritance law offers a highly informative benchmark for assessing the reasoning abilities of modern language models. 
To support research in this direction, recent work introduced MAWARITH \cite{bouchekif2026mawarith}, a large-scale benchmark of 12{,}500 Arabic inheritance cases designed for end-to-end reasoning from natural language descriptions. The benchmark includes detailed intermediate reasoning steps and final answers, making it possible to evaluate not only the correctness of the final output but also the validity of the reasoning process itself. This resource created an opportunity to move beyond standard multiple-choice evaluation and toward a more realistic setting in which systems must solve inheritance cases as humans would read and understand them.

The QIAS 2026 Shared Task was designed to evaluate whether participating systems can perform end-to-end Islamic inheritance reasoning in Arabic. It focuses on solving full cases from natural language, covering the complete reasoning process from heir identification to final share allocation. It also examines whether recent reasoning-oriented LLMs, such as Gemini, GPT, DeepSeek, Fanar, and Qwen, can transfer their strong performance on mathematical and synthetic benchmarks to the more complex domain of structured legal and religious reasoning.

In this paper, we present an overview of the QIAS 2026 Shared Task. We describe the task, the MAWARITH benchmark used in the evaluation, the MIR-E multi-step evaluation metric, the participating systems, and the main results obtained by the submitted approaches. We also discuss key lessons learned from the shared task and highlight the main challenges that current systems still face in Islamic inheritance reasoning. 

\section{Task Description}
The QIAS 2026 Shared Task focuses on the end-to-end automation of Islamic inheritance reasoning (\textit{ʿilm al-mawārīth}) from natural language. The task requires systems to process Arabic inheritance cases and generate a complete, structured solution. For each input question, systems must produce a detailed step-by-step reasoning trace (\texttt{<think>}), followed by a concise final answer (\texttt{<answer>}).

The task is formulated as a sequence of dependent reasoning stages. Systems are expected to identify and explicitly report the following components:
\begin{enumerate}[noitemsep, topsep=0pt]
    \item Identification of all mentioned heirs, including determining which are eligible or excluded based on applicable blocking rules (\textit{hajb}).
    \item Assignment of the correct legal shares (\textit{furūḍ}) to entitled heirs according to classical Islamic inheritance jurisprudence, following the majority opinion (\textit{al-jumhūr}).
    \item Determination of whether a global adjustment to the distribution is required.
    \item Computation of the final estate distribution, including cases where adjustments such as \textit{ʿawl} (proportional reduction) or \textit{radd} (return of the residue) apply.
\end{enumerate}

Participants must submit outputs in a structured format (e.g., JSON) that captures these reasoning steps. 
This representation enables fine-grained evaluation and allows analysis of different error types, such as legal reasoning errors versus numerical computation errors.

\section{Data}
\label{sec:data}
The QIAS 2026 data come from the MAWARITH benchmark introduced by \citet{bouchekif2026mawarith}. It contains 12{,}500 cases written in Arabic and follows the majority opinion in Islamic inheritance law (\textit{al-jumhūr}). Table \ref{tab:case_types} shows the distribution of cases across the training and test splits, as well as the complexity categories covered in the benchmark.
\begin{table}[!ht]
\centering
\begin{tabular}{l r r r r}
\hline
\textbf{Split} & \textbf{Simple } & \textbf{ʿAwl } & \textbf{Radd }  & \textbf{Total } \\
\hline
Training & $11{,}079$ & $577$   & $344$ &  $12{,}000$\\
Test     & $456$ & $39$ & $5$   & $500$\\
Total    & $11{,}535$ & $616$ & $349$  & $12{,}500$ \\
\hline
\end{tabular}
\caption{Distribution of inheritance cases by legal complexity}
\label{tab:case_types}
\end{table}
Each case describes a complete inheritance situation in natural language. The system must identify the heirs mentioned in the case, determine which heirs are excluded by blocking rules, assign the correct legal shares, decide whether an adjustment is needed, and compute the final distribution of the estate. In this way, the benchmark does not test only final answers, but the full reasoning process required to solve an inheritance case correctly. The dataset covers a wide range of family relations found in classical Islamic inheritance law, including parents, children, spouses, siblings, grandparents, uncles, nephews, and other extended relatives.
The \textsc{MAWARITH} dataset was built in several stages. First, inheritance cases were generated using the Almawarith calculator, which allows users to define heirs and their numbers through a structured interface and then produces the corresponding shares. This step provided a reliable foundation and helped ensure the correctness of the legal and numerical outcomes. Since the goal of \textsc{MAWARITH} is to evaluate reasoning from natural language, these structured cases were then rewritten as fluent Arabic inheritance questions, closer to real user queries. The outputs were then reviewed and enriched by an expert in Islamic studies, who added detailed legal and numerical explanations for each case. These explanations cover the main stages of inheritance reasoning, including heir identification, blocking rules, share assignment, and adjustment cases such as \textit{ʿawl} and \textit{radd} when needed. To improve consistency, the expert-written explanations were standardized with the support of Gemini-2.5-Flash, while keeping the legal reasoning unchanged. Finally, the dataset was carefully validated to ensure consistency between the question, the reasoning steps, and the final inheritance shares.

\subsection{MIR-E: Mawarith Inheritance Reasoning Evaluation}

\begin{table*}[!ht]
\centering
\setlength{\tabcolsep}{6pt}
\renewcommand{\arraystretch}{1.1}
\rowcolors{2}{gray!12}{white}
\begin{tabularx}{0.9\textwidth}{>{\raggedright\arraybackslash}p{0.45\textwidth} >{\raggedright\arraybackslash}X}
\toprule
\textbf{Team Name} & \textbf{Affiliations} \\
\midrule
CVPD \cite{swaileh2026cvpd} & CVPD Research group \\
Simplicity \cite{almansour2026simplicity} & King Saud University \\
KMS \cite{alkhamis2026kms}& King Saud University \\
QU-NLP \cite{alsmadi2026qu} & Qatar University \\
PSL \cite{mouhoub2026psl} & Paris Dauphine University  \\
Silah \cite{Ghader2026Silah} & Umm Al-Qura University \\
AGS-KSU \cite{Hicham2026agsksu} & King Saud University \\

\bottomrule
\end{tabularx}
\caption{Affiliations of teams that participated in the test phase and submitted a paper to QIAS 2026.}
\label{tab:teams_affiliations}
\end{table*}
The QIAS 2026 shared task uses MIR-E (Mawarith Inheritance Reasoning Evaluation) \cite{bouchekif2026mawarith}, a weighted multi-stage metric for evaluating both intermediate reasoning steps and final outputs in Islamic inheritance problems. Unlike standard evaluation based only on the final answer, MIR-E provides a more fine-grained assessment by scoring the main stages of the reasoning process. It includes four components:
\begin{enumerate}[noitemsep, topsep=0pt]
\item \textbf{Heirs and Blocking ($S_h$):} evaluates whether the model correctly identifies the effective heirs, the blocked heirs, and their counts.\item \textbf{Share Assignment ($S_s$):} measures whether the assigned shares for the eligible heirs are correct.\item \textbf{Adjustment ($S_a$):} checks whether the model predicts the correct adjustment type (\textit{none}, \textit{ʿawl}, or \textit{radd}), and is scored only if the previous two stages are fully correct.\item \textbf{Final Distribution ($S_f$):} evaluates whether the model produces the correct final distribution after the full inheritance process.\end{enumerate}

\section{Results and Discussion}
\begin{table*}[h]
\centering
\begin{tabular}{cllc}
\toprule
\textbf{Rank} & \textbf{Team} & \textbf{Approach} &\textbf{MIR-E}  \\
\midrule
1 & CVPD & Hybrid RAG &0.935  \\
2 & Simplicity  & Two-stage neuro-symbolic pipeline& 0.931  \\
3& KMS & Prompting-based end-to-end reasoning &0.916 \\
4 & QU-NLP & Multi-stage QLoRA fine-tuning &0.907 \\
5 & PSL & prompting-based end-to-end reasoning &0.898  \\
6   & AGS-KSU & QLoRA fine-tuning and Prompting-based end-to-end reasoning & 0.84  \\
7 & Silah & LoRA fine-tuning &  0.826  \\
8 & UTLM  & Prompting-based end-to-end reasoning & 0.742  \\
\bottomrule
\end{tabular}
\caption{Official leaderboard results for QIAS 2026 for teams that participated in the test phase and submitted a paper}
\label{tab:leaderboard}
\end{table*}
A total of 16 teams participated in the final phase. Table 2 summarizes the affiliations of the teams that participated in the test phase and submitted a paper. We provide a baseline implementation using Fanar-Sadiq, a modern Arabic large
language model accessible via API. This baseline
relies exclusively on prompting techniques, without
any fine-tuning. The goal is to provide a simple
yet effective reference point for evaluating model
performance. The dataset and baseline code are publicly available online.\footnote{\url{https://gitlab.com/islamgpt1/qias_shared_task_2026}} Overall, the submitted systems illustrate three main methodological directions: \textit{(i)} end-to-end reasoning with prompting alone, \textit{(ii)} fine-tuned models specialized for the task, and \textit{(iii)} hybrid pipelines that combine LLM-based language understanding with deterministic symbolic reasoning. This diversity of approaches makes the shared task especially useful for comparing different strategies for Arabic legal reasoning. The participating systems explored different ways of solving Islamic inheritance cases. 
The most common approach explored by participants was prompting-based end-to-end reasoning with large language models. Team PSL  \cite{mouhoub2026psl}  followed this setting and evaluated several models, including Gemini 2.5 Flash, Qwen3-32B, GPT-oss-120B, Llama-3.3-70B, Fanar-Sadiq, and Fanar-C-2-27B. Team KMS \cite{alkhamis2026kms} explored a similar direction and also evaluated Gemini 2.5 Pro and Mistral. Overall, the two teams reported similar performance trends: commercial models were generally more reliable, while  open-weight models  showed weaker results on this task.

Team CVPD \cite{swaileh2026cvpd}, which achieved the best result in the shared task, proposed a RAG-based pipeline designed to generate outputs that match the MIR-E evaluation format. Their system built a knowledge base from synthetic legal question-answer pairs, with the option to include books, then retrieved relevant context for each Arabic inheritance question before generating a single JSON output. This output followed the required schema and included eligible heirs, blocked heirs, legal shares, the adjustment type when applicable, and the final \textit{taṣīl} distribution. The system also included parsing and validation steps to ensure compatibility with the official evaluator. This approach showed that combining retrieval, controlled generation, and structured output constraints can be highly effective for this task.

Team Silah  \cite{Ghader2026Silah} investigated three strategies: retrieval-augmented generation based on a curated rule base, supervised fine-tuning of large language models, and a combination of fine-tuning and retrieval. Their experiments showed that fine-tuning alone performed better than retrieval-based approaches, with the best results obtained by a fine-tuned Fanar model. This suggests that task-specific fine-tuning can be an effective approach for Islamic inheritance reasoning. Team QU-NLP \cite{alsmadi2026qu} proposed a multi-stage Quantized Low-Rank Adaptation (QLoRA) strategy. Their approach involved initial domain adaptation on a corpus of Islamic fatwas to capture jurisprudential reasoning patterns, followed by task-specific fine-tuning on structured inheritance cases to optimize JSON-formatted output. This methodology allowed a relatively small 4B parameter model to achieve competitive performance, highlighting the effectiveness of specialized training strategies for complex legal reasoning tasks. Team AGS-KSU \cite{Hicham2026agsksu} also explored a fine-tuning approach using Qwen2.5-3B, similar to the strategy adopted by Team QU-NLP. However, this fine-tuned model achieved a relatively low MIR-E score of $0.30$. By comparison, their prompting-based configuration using GPT-5.4 Thinking reached 0.84, yielding a substantially stronger result.

Finally, some teams explicitly separated natural-language understanding from legal computation. Team Simplicity \cite{almansour2026simplicity}  proposed a two-stage neuro-symbolic pipeline in which a commercial LLM was used only for Arabic information extraction. The extracted heirs were mapped to a standardized set of legal heir categories and then passed to a symbolic rule-based component that carried out blocking, share assignment, and final calculation according to the rules of \textit{ʿilm al-mawārīth}. This design reflects a clear division of labor between the LLM and the symbolic module.

The final performance of the submitted systems is summarized in Table \ref{tab:leaderboard}, where Team CVPD achieved the top ranking (0.935) using a RAG pipeline based on the Qwen-9B model. Notably, the QU-NLP system’s multi-stage QLoRA approach achieved a competitive MIR-E score of 0.907. This performance is particularly significant, as it closely matches the 0.901 score of the commercial Gemini-2.5-Flash model (as reported in \cite{bouchekif2026mawarith}). These results demonstrate that, while RAG remains a powerful tool for precision, specialized domain adaptation can enable smaller open-weight models, such as the 4B-parameter Qwen3, to achieve performance levels close to those of commercial models on this task.

\section{Conclusions and Future Work}
In this paper, we presented the QIAS 2026 Shared Task, designed to evaluate the ability of large language models to solve Islamic inheritance cases end-to-end, including case understanding, heir identification, rule application, and final share distribution. The results of the participating systems revealed a clear gap between commercial and open-weight models, with commercial systems generally achieving stronger performance on this challenging reasoning task. At the same time, the findings suggest that fine-tuning LLMs on MAWARITH data can substantially improve performance, while RAG also proved beneficial for some systems.

As a future direction, we plan to organize a follow-up edition of the shared task focusing on more complex inheritance cases, including pregnancy-related cases, multiple deaths, the missing person (mafqūd), and the intersex heir (khunthā). We also plan to focus on fine-tuning approaches for small Arabic LLMs. To ensure a fairer comparison across participants, future editions may restrict the use of commercial LLMs and focus on open-weight models that can be reproduced and compared under the same conditions.

\section{Limitations}

This work has several limitations. First, QIAS 2026 focuses only on inheritance cases based on the majority opinion (\textit{al-jumh\={u}r}). As a result, it does not represent the full diversity of juristic opinions in Islamic law.
Second, although the benchmark covers many cases, it still does not include all the complexity of real inheritance situations. Some difficult or rare cases remain outside the current scope.

Third, the MIR-E metric evaluates structured reasoning outputs in a detailed and reproducible way, but it does not fully measure explanation quality, clarity, or usefulness for real users.

Finally, some submitted systems use commercial models, which makes full reproducibility more difficult. Future work should expand the legal coverage of the benchmark and give more attention to open and reproducible systems.


\section{References}\label{sec:reference}
\bibliographystyle{lrec2026-natbib}
\bibliography{lrec2026-example}

\label{lr:ref}
\bibliographystylelanguageresource{lrec2026-natbib}
\bibliographylanguageresource{languageresource}

\appendix
\section{Related Work} \label{rl}
Large language models have recently been applied to a wide range of Islamic knowledge tasks, including Quranic question answering~\cite{bhatia2026ragagenticragfaithful}, 
knowledge retrieval~\cite{mubarak-etal-2025-islamiceval, xuan-phuc-dang-van-2025-puxai}, and the analysis of hallucinations in Islamic content~\cite{mubarak-etal-2025-islamiceval}.
A recent related benchmark is \textit{IslamicMMLU} \cite{abdelaal2026islamicmmlubenchmarkevaluatingllms}, which evaluates LLMs on broad Islamic knowledge across the Quran, Hadith, and Fiqh through a large multiple-choice benchmark.
These studies show that LLMs perform well on knowledge retrieval and basic understanding when answers rely on direct textual matching. However, they often hallucinate and show clear limitations on tasks that require structured reasoning or deep domain knowledge.
In \cite{bouchekif2025islamic}, the authors report that several models, such as LLaMA and ALLaM, frequently cite non-existent Quranic verses or fabricate Hadith references.
As a result, the generated conclusions are not only incorrect, 
but are also supported by false religious evidence. 
This behavior raises serious concerns for religious and legal applications, where correctness depends not only on the final answer but also on the authenticity and reliability of the cited sources.

To mitigate hallucinations, recent work has explored Retrieval-Augmented Generation (RAG) approaches. While RAG improves access to relevant information and enhances factual faithfulness 
and citation accuracy~\cite{bhatia2026ragagenticragfaithful, noureldien-etal-2025-athar, alowaidi-2025-sea}, it remains insufficient for answering questions that require multi-step reasoning. This limitation has motivated the development of reasoning-oriented models that explicitly support multi-step inference. In this context, recent research has increasingly focused on such models that aim to move beyond surface-level text generation and promote more reliable reasoning. 

Models such as \textit{o3}, \textit{GPT-5} \cite{singh2025openai}, \textit{Gemini-2.5}~\citep{team2023gemini}, \textit{Gemini3}, \textit{DeepSeek-R1}~\citep{deepseek2024r1},  along with open models such as \textit{Fanar-C-2-27B}~\cite{abbas2026fanar}, \textit{Falcon-H1R}~\cite{chaabane2026falcon}, \textit{Fanar-Sadiq} \cite{abbas2026fanar} and \textit{Qwen3} ~\citep{shen2025qwenlongl15posttrainingrecipelongcontext} illustrate this trend by promoting more consistent multi-step inference through instruction tuning and large-scale pretraining. 
Evaluations of reasoning-oriented language models have largely focused on mathematical and logical benchmarks, on which these models have achieved strong results, particularly in arithmetic reasoning, symbolic manipulation, and competition-style mathematics~\cite{abs-2110-14168, hendrycks2021measuringmathematicalproblemsolving, wei2023chainofthoughtpromptingelicitsreasoning}. 
Beyond mathematical and logical benchmarks, recent work has begun to investigate LLM reasoning in legally grounded settings by evaluating models on legal benchmarks such as BRIEFME~\cite{woo2025briefme}, which require structured argumentation and rule-based reasoning.
The authors show that \textit{GPT-4o} can outperform human annotators on argument summarization by producing clear and coherent summaries.
Even within the Islamic domain, inheritance law has received growing attention as a challenging testbed for LLM reasoning \cite{aldahoul-zaki-2025-nyuad-qias, rbaiti-etal-2025-morai, farouk-zaki-2025-cis}.
In particular, QIAS 2025\footnote{https://sites.google.com/view/qias2025/} \cite{bouchekif-etal-2025-qias} was a shared task dedicated to Islamic inheritance law (\textit{ʿilm al-mawārīth}), focusing on the evaluation of large language models under strict, rule-based legal and numerical constraints, using a benchmark of $2,200$ MCQs.  A similar MCQs benchmark is \textit{MirathQA} \cite{ALMASOUD2026112589}, built from $1394$ inheritance cases.  Studies report that commercially deployed (\textit{e.g., Gemini and ChatGPT}), reasoning-oriented models consistently outperform non-reasoning or general-purpose models on benchmarks requiring multi-step inference and structured reasoning ~\cite{bouchekif2025islamic,al-smadi-2025-qu,bekhouche2025cvpd, motasim-hamed-etal-2025-hiast,hossain-afli-2025-adapt-mtu}. Additionally,  ~\citet{elrefai-etal-2025-gumball} show that a fine-tuned Qwen3 model achieved top-ranked performance on the   QIAS 2025 shared task. However, this evaluation setup does not allow assessing whether models truly reason correctly. Models were required to select a single correct answer among six options, without any evaluation of the validity of their intermediate reasoning steps or the correctness of the legal justifications leading to that choice. Moreover, \citet{bouchekif2025islamic} shows that even when a model selects the correct answer, the underlying reasoning can still be incorrect or legally invalid. In contrast, the present shared task requires models to perform end-to-end inheritance reasoning, explicitly generating intermediate reasoning steps, applying jurisprudential rules, and computing the final inheritance shares.

\end{document}